\documentclass[letterpaper, 10 pt, conference]{ieeeconf}  

\IEEEoverridecommandlockouts                              

\usepackage[T1]{fontenc}
\usepackage{marvosym}
\usepackage[utf8]{inputenc}
\usepackage{graphicx}
\usepackage{hyperref}
\usepackage{CJKutf8}
\usepackage{float}
\usepackage{svg} 
\usepackage{pdfpages}
\usepackage{subfigure}
\usepackage{tabularx}
\usepackage{booktabs}
\usepackage{amsmath}
\usepackage{amssymb}
\usepackage{mathtools}
\usepackage{bm}
\usepackage{amssymb} 
\usepackage{CJKutf8}
\usepackage{array}
\usepackage{booktabs}
\usepackage{multirow}
\usepackage{multicol}
\usepackage{color, colortbl}
\usepackage{pifont}%
\usepackage{array}
\usepackage{makecell}
\usepackage{balance}
\usepackage{tabularray}
\usepackage{tabularx, booktabs}

\usepackage{amsmath}
\usepackage{algorithm}
\usepackage{algorithmic}

\usepackage{caption}
\usepackage{booktabs}

\newcolumntype{Y}{>{\centering\arraybackslash}X}
\title{\LARGE \bf
AccidentBlip: Agent of Accident Warning based on MA-former
}

\author{Yihua Shao\textsuperscript{1,2}, Yeling Xu\textsuperscript{3}, Xinwei Long\textsuperscript{3}, Siyu Chen\textsuperscript{1}, Ziyang Yan\textsuperscript{3},
\\Yang Yang\textsuperscript{1\Letter}, Haoting Liu\textsuperscript{4}, Yan Wang\textsuperscript{3}, Hao Tang\textsuperscript{2}, Zhen Lei\textsuperscript{1}
\thanks{$^{1}$Institute of Automation, Chinese Academy of Sciences, Beijing, China. $^{2}$Peking University, Beijing, China. 
 $^{3}$Tsinghua University, Beijing, China. $^{4}$University of Science and Technology Beijing，Beijing, China.
}%
\thanks{\raggedright{Corresponding authors:{\tt\small \quad yang.yang@nlpr.ia.ac.cn} } }
}

\begin{document}

\begin{CJK}{UTF8}{gbsn}

\maketitle
\thispagestyle{empty}
\pagestyle{empty}

\begin{abstract}


In complex transportation systems, accurately sensing the surrounding environment and predicting the risk of potential accidents is crucial. Most existing accident prediction methods are based on temporal neural networks, such as RNN and LSTM. Recent multimodal fusion approaches improve vehicle localization through 3D target detection and assess potential risks by calculating inter-vehicle distances. However, these temporal networks and multimodal fusion methods suffer from limited detection robustness and high economic costs.
To address these challenges, we propose AccidentBlip, a vision-only framework that employs our self-designed Motion Accident Transformer (MA-former) to process each frame of video. Unlike conventional self-attention mechanisms, MA-former replaces Q-former's self-attention with temporal attention, allowing the query corresponding to the previous frame to generate the query input for the next frame. Additionally, we introduce a residual module connection between queries of consecutive frames to enhance the model's temporal processing capabilities.
For complex V2V and V2X scenarios, AccidentBlip adapts by concatenating queries from multiple cameras, effectively capturing spatial and temporal relationships. In particular, AccidentBlip achieves SOTA performance in both accident detection and prediction tasks on the DeepAccident dataset. It also outperforms current SOTA methods in V2V and V2X scenarios, demonstrating a superior capability to understand complex real-world environments.
\end{abstract}

\section{INTRODUCTION}


The prevention and detection of traffic accidents is a critical issue in the field of intelligent transportation \cite{world2019global, ren2022tbsm}. Traditional approaches \cite{guo2022responsibility, yu2022vehicle} often employ temporal networks, such as LSTM \cite{hochreiter1997long, yu2020identifying} and RNN \cite{wang2019state}, to process temporal information in driving scenarios \cite{lee2017real}. These methods typically use a classifier to detect or predict the occurrence of an accident \cite{worrall2010improving}.
Some studies have also explored scenario-based simulations to emulate autonomous driving environments for initial model testing \cite{mokhtari2021don}. However, these traditional methods have limitations: they lack compatibility with end-to-end autonomous driving systems and depend heavily on prior knowledge of the surrounding environment for accurate detection.


With the advent of Transformers \cite{shao2024gwq}, question-and-answer tasks are increasingly replacing traditional detection and classification networks. Models like BERT \cite{devlin2018bert} utilize extensive pre-training on large datasets to facilitate textual querying. As multimodal models continue to evolve, visual-textual pre-trained models have emerged, enabling the detection and description of individual images \cite{couto2018introduction}.
For instance, Blip \cite{li2022blip} and Blip2 \cite{li2023blip} are pre-trained on image-text pairs, supporting tasks such as image-based Q\&A and description. Notably, Blip2's Q-Former is frequently integrated with autonomous driving models \cite{jin2025tod3cap}. Building on this, models such as Video-LLaMA \cite{zhang2023video} and Video-Vicuna \cite{chiang2023vicuna} extend Q-Former's capabilities to process temporal input frames, aggregating them to generate video-based Q\&A outputs. Similarly, Video-LLaVA \cite{lin2023video} incorporates frame-to-text projections, enhancing long-video comprehension and improving accuracy.
However, these models lack support for multi-view inputs, which limits their applicability in autonomous driving scenarios. In multi-vehicle environments, V2Xformer \cite{wang2024deepaccident} addresses accident detection through multimodal fusion techniques \cite{liu2023bevfusion}. Nevertheless, its reliance on LiDAR-vision fusion introduces significant memory overhead, making practical deployment on vehicles challenging.



To address these limitations, we introduce AccidentBlip, a Transformer-based, multi-view accident warning agent designed for Q\&A tasks on temporal traffic images. AccidentBlip can perform both accident detection and prediction in complex traffic scenarios. At its core, AccidentBlip incorporates our self-designed Motion Accident Transformer (MA-former), which is specifically designed to process temporal information.
Unlike existing accident detection approaches, AccidentBlip avoids the use of large and bloated language models \cite{wang2024accidentgpt} or computationally expensive multimodal fusion methods \cite{wang2024deepaccident}. Instead, it relies solely on visual reasoning, making it more efficient and practical for autonomous driving applications.
The MA-former processes temporal information by replacing the self-attention mechanism of Blip2’s Q-Former \cite{li2023blip} with temporal attention. This design allows the MA-former to accept query outputs from the previous frame and generate updated queries for the subsequent frame, thereby enabling effective temporal information processing.


\begin{figure*}
    \centering
    \includegraphics[width=1.0\linewidth]{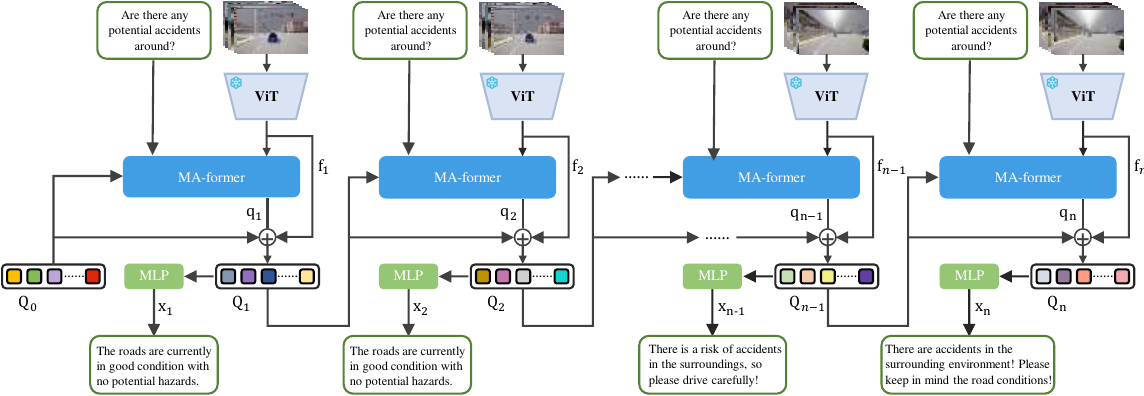}
    \caption{\textbf{Overall architecture of AccidentBlip.} We input the multi-view images of each frame into ViT to obtain the features of each frame. Then we input previous frame query and multi-view images features into MA-former and input the text description. Finally, we use residual to connect the previous frame query and the image features of this frame with MA-former's output and input it into the MA-former corresponding to the next frame.}
    \label{fig:pipeline}
    \vspace{-0.4cm}
\end{figure*}

In addition to enabling the model to process temporal multi-view images, we evaluated AccidentBlip's performance in multi-vehicle systems and V2X scenarios. AccidentBlip outperforms existing accident detection and prediction models on the DeepAccident dataset \cite{wang2024deepaccident}.
In summary, the contributions of this paper are as follows:

\begin{itemize}
  \item [1)] 
    We introduce AccidentBlip, a Transformer-based accident warning agent that utilizes vision-only inference to detect and predict accidents surrounding the ego vehicle.
  \item [2)]
    We propose MA-former, a novel module that processes temporal multi-view images by replacing Q-Former's self-attention mechanism with temporal attention.
    \item [3)] 
    AccidentBlip achieves state-of-the-art performance on the DeepAccident dataset, demonstrating high recall precision and average precision for the accident detection task, as well as superior accuracy for the accident prediction task.
\end{itemize}

\section{RELATED WORK}

\subsection{End-to-End Automatic Driving}


Recently, with the rapid development of autonomous driving models \cite{yan2024renderworld, remondino2023critical, yang2023survey, yan20243dsceneeditor, yan2023nerfbk}, end-to-end autonomous driving is increasingly emerging in the fields of intelligent driving and complex traffic perception. A notable example is UniAD \cite{hu2023planning}, which integrates the three critical components of autonomous driving—perception, decision-making, and planning—into a unified network architecture.
In the perception task, GPT-DRIVER \cite{mao2023gpt} employs text boxes to detect common objects, such as vehicles, in each frame and sends the resulting bounding box coordinates to ChatGPT as linguistic information. Similarly, models like BEVFormer \cite{li2022bevformer, yang2023bevformer} utilize 3D target detection techniques \cite{wang2022detr3d} for road objects to enable end-to-end perception. Another example is DriveGPT4 \cite{xu2024drivegpt4}, which proposes a Q\&A-based approach for road situation perception. However, this model focuses solely on front-view inputs and evaluates performance on the BDD-X dataset \cite{kim2018textual}, overlooking multi-view scenarios crucial in real-world driving.
Despite their advancements, most of these methods are limited to the measurement stage and face challenges in enabling V2V or V2X collaborative sensing, which are essential for broader traffic awareness and coordination.

\subsection{Accident Detection and Prediction}

Traffic accident detection and prediction is a critical research area in traffic safety \cite{world2019global, ren2022tbsm}. Traditional detection methods \cite{guo2022responsibility, yu2022vehicle, hochreiter1997long, yu2020identifying} typically utilize the vehicle's front view in combination with neural networks for temporal detection \cite{wang2019state}. However, these approaches often fail to adequately sense dangers in complex traffic environments.
With the advent of Transformer architectures \cite{vaswani2017attention}, Transformer-based methods such as AccNet \cite{liao2024real} leverage temporal attention to analyze sequential traffic scenes. Despite their promise, most of these models \cite{karim2022toward, wang2023gsc} are tested on datasets such as DADA-2000 \cite{fang2019dada}, which only include ego-vehicle front-view data. This limitation makes it challenging to apply these models to complex transportation systems with V2X scenarios.
The emergence of video-language models (Video LLMs) \cite{zhang2023video, lin2023video} has also sparked interest in applying them to accident detection tasks. However, the lack of multi-view input capabilities hinders their deployment in autonomous driving vehicles \cite{zhang2023video, long2024generative}.
To address these challenges, V2Xformer \cite{wang2024deepaccident} was the first to propose utilizing multimodal fusion \cite{liu2023bevfusion} and 3D target detection \cite{wang2022detr3d} for accident prediction. Although effective, reliance on expensive LiDAR technology significantly increases costs. Building on this, AccidentGPT \cite{wang2024accidentgpt} introduces advanced reasoning capabilities based on V2Xformer. However, its large training parameter size results in a much higher operational cost compared to other methods.

\section{METHODOLOGY}

In this section, we introduce the overall framework of our model, and its adaptation to complex V2X and V2V traffic scenarios.

\subsection{Overall Architecture }
\label{arch}

As shown in Fig. \ref{fig:pipeline}, we first input the zero initialized query $Q_0$ into the MA-former corresponding to the first frame and process each multi-view frame with ViT-14g pretrained by EVA-CLIP \cite{sun2023eva}. In order to input the multi-view features output from ViT-14g into the MA-former, we flatten and concatenate frame features into a fusion feature $f_n$. Therefore, query $q_n$ of each frame output from MA-former can be represented as Eq. \eqref{qn}.
\begin{equation}
   q_n=\operatorname{MA-former}\left (  Q_{n-1},f_{n}  \right ) ,
   \label{qn}
\end{equation}
where MA-former we will explain in Sec. \ref{MA-former}.

To enhance and preserve more features between consecutive frames, we employ a residual connection \cite{he2016deep} that connects frame $n$'s multi-view feature $f_n$, MA-former output query $q_n$ with previous frame input query $Q_{n-1}$. The process can be represented as Eq. \eqref{Qn}.
\begin{equation}
    Q_n = Q_{n-1} \oplus \text{MA-former}(Q_{n-1}, f_n) \oplus f_n,
    \label{Qn}
\end{equation}
where $Q_n$ is the query used for text output of frame $n$.
The residual output query $Q_n$ is fed into an MLP for binary classification, allowing us to determine the risk of an accident near the ego vehicle. This process can be represented as Eq.~\eqref{result}.
\begin{equation}
    Result_{n} =\operatorname{MLP}\left ( Q_{n}  \right )  ,
    \label{result}
\end{equation}
where $Result_{n}$ represents the result of the frame $n$.
To enable the agent to predict accident risk, the token output from the MLP is input into embedding $x_n$ to describe the scenario.

\subsection{The Proposed MA-former}
\label{MA-former}

\begin{figure}
    \centering
    \includegraphics[width=1.0\linewidth]{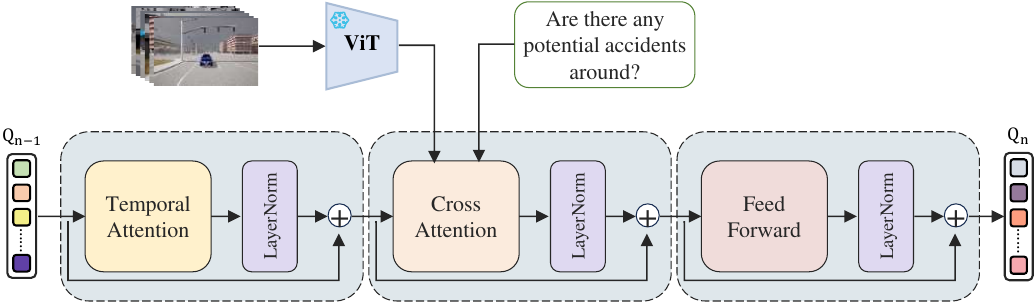}
    \caption{\textbf{Architecture of MA-former.} MA-former replaces Q-former's self-attention with temporal attention to realize processing temporal images.}
    \label{fig:MA-former}
    \vspace{-0.4cm}
\end{figure}






In order to process the information of the past scene, we design the MA-former to process the query corresponding to the previous frame. As shown in Fig. \ref{fig:MA-former}, we utilize the temporal attention of MA-former to receive the previous frame's query. The process can be represented as Eq. \eqref{qQ'}.
\begin{equation}
     Q_{n-1}'=\operatorname{TemporalAttn}(Q_{n-1},K_{n}, V_n),
     \label{qQ'}
\end{equation}
so the output of temporal attention can be denoted as Eq. \eqref{QKV}.
\begin{equation}
    Q_{n-1}'= \operatorname{Softmax}\left(\frac{Q_{n-1} \cdot K_{n}^T}{\sqrt{d_k}}\right)V_n,
    \label{QKV}
\end{equation}
where $Q_{n-1}'$ is the output of temporal attention, $Q_{n-1}$ is the query corresponding to the $n-1$ frame, $K_n$ and $V_n$ are the key and value matrices of temporal attention for the current frame, $d_k$ is the dimension of the K-matrix.

To realize multimodal input, we utilize the output of the temporal attention query $Q_{n-1}$ with the multi-view image feature $f_n$, text query $q'$ to compute cross attention. This process can be represented as Eq. \eqref{qk} and \eqref{q''}.
\begin{equation}
    Q = (Q_{n-1}'+q_n) W^Q, \quad K = V = f_n W^K,
    \label{qk}
\end{equation}
\begin{equation}
    Q''= \operatorname{CrossAttn}(Q_{n-1}',q', f_n),
    \label{q''}
\end{equation}
so the cross attention can be represented as Eq. \eqref{Q2}.
\begin{equation}
    Q'' = \operatorname{Softmax}\left(\frac{Q \cdot K^T}{\sqrt{d_k}}\right)V ,
    \label{Q2}
\end{equation}
where $Q''$ is the output of cross attention, $Q_{n-1}'$ is the output query from $Q_{n-1}$ temporal attention stage. $Q$ is the projection mapping from the feature $Q_{n-1}'$ and $q'$, $K$ and $V$ is the features derived from $f_n$.

We then input $Q''$ into the feedforward for the high-level query $q_n$ and input it into the residuals connection which is presented as Eq. \eqref{Qn}. From this we can obtain the output text query $Q_n$ corresponding to the frame $n$.

We successfully implement modeling the temporal scenes and traffic state detection based on ego surroundings using visual text input by using MA-former.

\subsection{Complex Systems Adaptation}

In complex transportation systems, applying the model to V2X or V2V scenarios requires effective multi-vehicle information fusion, which we achieve through multi-query aggregation.

In our approach, we utilize ViT-14g to extract features $f$ from multi-view images of each vehicle in the current frame. Subsequently, these features are processed by the MA-former. For each vehicle, the queries are fused according to their timestamps to integrate information from different viewpoints. The fusion process is described by the Eq.\eqref{q_fused}:
\begin{equation}
    Q_{used}^n = \operatorname{Linear}(\operatorname{concat}(q_{c1}, q_{c2}, q_{c3}, \cdots, q_{ci})),
    \label{q_fused}
\end{equation}
where $q_{ci}$ represents the query from the $i$-th car's viewpoint, and $Q_{fused}^n$ denotes the fused query for the $n$-th frame.

After that, the fused query for each frame is input into an MLP. Similar to the single vehicle case, MLP outputs whether there is an accident risk in the scenario based on the input fused query $Q_{fused}^n$. The process can be described as Eq.~\eqref{r_m}:
\begin{equation}
    Result_{fused}^n = \operatorname{MLP}(Q_{fused}^n).
    \label{r_m}
\end{equation}

We input the output of the MLP into the embedding $X_n$ and then get the current state of the scene. This approach ensures the effective fusion of multi-view and multi-vehicle information, enabling accurate accident prediction and better applicability to V2X or V2V systems.

\begin{figure}
    \centering
    \includegraphics[width=1\linewidth]{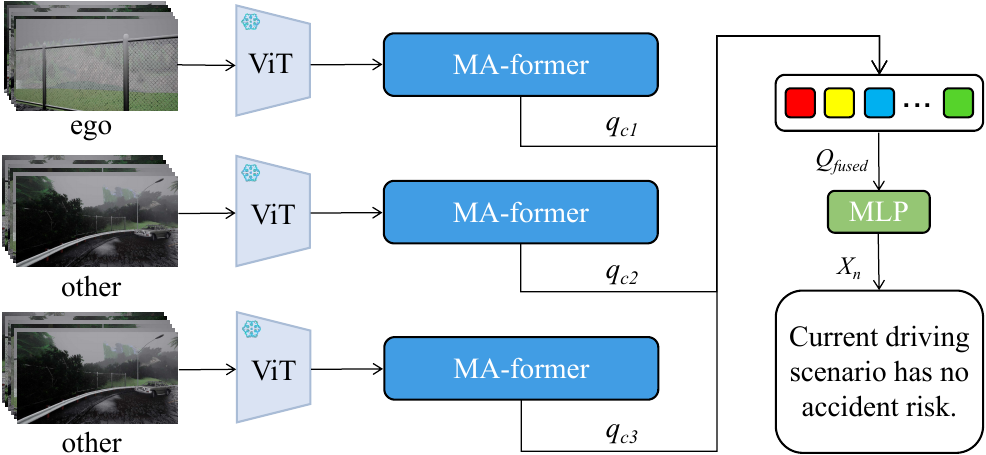}
    \caption{\textbf{Architecture of multi-vehicle sensing system. }We realize the fusion of multi-vehicle information by merging the queries output from each on-vehicle MA-former.}
    \label{fig:enter-label}
    \vspace{-0.4cm}
\end{figure}

\section{EXPERIMENTS}

\begin{table}[!t]
    \centering
    \begin{tabularx}{245pt}{c|c*{6}{Y}}
    \toprule
    &\multicolumn{3}{c}{\textbf{ego}}  & \multicolumn{3}{c}{\textbf{ego + infra}}\\
    \textbf{Method}& Pre.  & Rec.  & APA& Pre.  & Rec.  & APA \\\midrule
    AccNet \cite{liao2024real}& 10.3 & 30.7 & 20.8& 11.7  & 30.8 & 21.6 \\
    Blip2 \cite{li2023blip}& 9.5 & 70.7 & 11.0& 9.1  & \textbf{76.8} & 9.4 \\
    Video-Llama \cite{zhang2023video}& 20.1 & \textbf{79.4} & 42.0  & 18.9  & 71.6 & 36.4  \\
    Video-Vicuna \cite{chiang2023vicuna}& 19.9  & 77.6 & 40.5 & 18.1  & 75.8 & 35.1  \\
    Video-LLaVA \cite{lin2023video}&30.4 & 78.8 & 45.7& 29.7  & 67.2 & 41.4 \\
    V2Xformer \cite{wang2024deepaccident}&54.8 & 45.6 & 61.9& 58.2  & 50.0 & 68.1 \\
    AccidentGPT \cite{wang2024accidentgpt}&59.3 & 62.8 & 62.6& 60.4  & 63.7 & 69.9 \\
    AccidentBlip (Our) &\textbf{62.6} & 68.4& \textbf{64.7} & \textbf{64.2}  & 70.5 & \textbf{72.4} \\ 
  \bottomrule
    \end{tabularx}
    \caption{\textbf{Ego and V2X accident detection results. }AccidentBlip reaches \textbf{SOTA} in accident detection precision and accident prediction accuracy.}
    \label{ego}
    \vspace{-0.4cm}
\end{table}

In this section, we evaluate our model's performance in accident detection and prediction tasks across V2X and V2V systems of varying complexity, demonstrating its potential for application in complex transportation systems. All experiments are conducted on NVIDIA A100 GPUs (80GB).

\subsection{Experiment Setting}

\noindent\textbf{Dataset.} We use the DeepAccident dataset \cite{wang2024accidentgpt} for model training and evaluation. The dataset contains 691 scenarios with 285K samples, split into training, validation, and test sets at ratios of 70\%, 15\%, and 15\%, respectively. Accidents and non-accidents account for 26.9\% and 73.1\% of the data. A one-meter threshold is used to evaluate accident prediction accuracy, with each scenario including data from four vehicles and one infrastructure node to facilitate V2X research.

\noindent\textbf{Baselines.} We select state-of-the-art (SOTA) methods as baselines, including the traditional AccNet \cite{liao2024real}, frame-by-frame Q\&A SOTA Blip2 \cite{li2023blip}, popular video understanding models Video-LLaMA \cite{zhang2023video}, Video-Vicuna \cite{chiang2023vicuna}, and Video-LLaVA \cite{lin2023video}, as well as multimodal fusion methods V2Xformer \cite{wang2024deepaccident} and AccidentGPT \cite{wang2024accidentgpt}.

\noindent\textbf{Training details.} During training, we use the Adam optimizer \cite{kingma2014adam} with default hyperparameters $\beta_1 = 0.9$ and $\beta_2 = 0.999$ and set the initial learning rate $lr=1 \times 10^{-5}$. A learning rate warm-up strategy is applied during the first three training epochs to mitigate instability at the start of training. After the warm-up phase, we adopt a cosine annealing learning rate scheduler to smoothly decrease the learning rate over the subsequent five epochs, promoting better convergence. The model is trained for a total of 8 epochs, with a batch size of 8. 

The input to the model consists of multi-view temporal data, represented as an input tensor with the following shape:
\begin{equation}
X_S \in \mathbb{R}^{T \times N_V \times V_C \times C \times H \times W},
\end{equation}
where $T$ represents the temporal length of the sequence, $N_V$ is the number of vehicles, and $V_C = 6$ is the number of viewpoints for each vehicle. The viewpoints include Front, FrontLeft, FrontRight, Back, BackLeft, and BackRight; $H$ and $W$ denote the height and width of the input image, while $C$ represents the number of color channels (typically 3 for RGB images). During the data preprocessing stage, all images are resized to $224 \times 224$ to meet the input requirements of the visual feature extraction module.

The training objective for the multi-vehicle system is to minimize the cross-entropy loss for accident classification, optimizing the model's ability to predict accidents in complex scenarios. The loss function $\mathcal{L}$ is defined as Eq. \eqref{loss}:
\begin{equation}
    \mathcal{L} = -\frac{1}{N} \sum_{i=1}^N \sum_{j=1}^C y_{ij} \log(p_{ij}),
    \label{loss}
\end{equation}
where $N$ is the number of samples in the batch, $C$ is the number of classes (2 in this task), $y_{ij}$ is the one-hot encoded true label, and $p_{ij}$ represents the model's predicted probability for class $j$. To adapt to this loss function, the model outputs logits without applying a softmax activation, as the loss function handles the required normalization internally.

\noindent\textbf{Evaluation Metrics.} We define the accident detection task as a binary classification task. For the binary classification task, we define the test results as true positive ($TP$), false positive ($FP$), true negative ($TN$) and false negative ($FN$). We propose precision ($\textbf{Pre}\uparrow$), recall ($\textbf{Rec}\uparrow$) and average precision ($AP\uparrow$) of the model as evaluation metrics. The precision and recall can be calculated by Eq. \eqref{PR}.
\begin{equation}
    \textbf{Pre}=\frac{TP}{TP+FP},~\textbf{Rec}=\frac{TP}{TP+FN},  \label{PR}
\end{equation}

For the accident prediction task, we use the benchmark Accident Prediction Accuracy ($\mathbf{APA}\uparrow$) proposed by DeepAccident \cite{wang2024deepaccident} to evaluate the model's capabilities. We calculate $\mathbf{APA}$ using Eq. \eqref{APA},
\begin{equation}
    \mathbf{APA} =\frac{1}{D}  \sum_{d\in D}^{} \frac{\left | TP_p \right |_{d}  }{\left | TP_p \right |_{d}+  \frac{1}{2}\left | FP_p \right |_{d} +  \frac{1}{2}\left | FN_p \right |_{d}}. 
    \label{APA}
\end{equation}
where $d$ refers to a specific threshold value from the set $D=\left \{5,10,15 \right \} $, which is used to evaluate the model's performance across different levels of strictness. $|TP_p|_d$, $|FP_p|_d$, and $|FN_p|_d$ are the counts of true positives, false positives, and false negatives at $d$ in the prediction task.

\begin{figure}[t]
    \centering
    \includegraphics[width=1\linewidth]{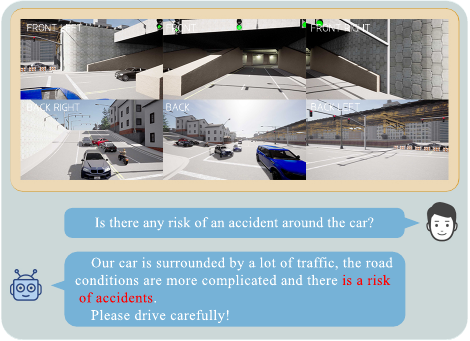}
    \caption{\textbf{Example of interaction between AccidentBlip and driver. }AccidentBlip assesses the risk of accidents in the surrounding environment and alerts the driver to drive with caution.}
    \label{fig:example}
    \vspace{-0.4cm}
\end{figure}

\subsection{Main Results}

\begin{table*}[h]
    \centering
    \begin{tabularx}{\textwidth}{c|c*{12}{Y}}
    \toprule
    &\multicolumn{3}{c}{\textbf{ego + behind vehicle}}  & \multicolumn{3}{c}{\textbf{ego + other vehicle}}&\multicolumn{3}{c}{\textbf{ego + behind + other vehicle}}  & \multicolumn{3}{c}{\textbf{4 vehicles}}\\
    \textbf{Method}& Pre.  & Rec.  & APA& Pre.  & Rec.  & APA & Pre.  & Rec.  & APA& Pre.  & Rec.  & APA\\\midrule
    AccNet \cite{liao2024real}&11.8 & 31.4 & 30.3&  12.7 & 32.0 & 28.9 & 12.1  & 33.7 & 28.0  & 11.9  & 34.8 & 27.4\\
    Blip2 \cite{li2023blip}& 10.8  & \textbf{66.4} & 11.6  & 12.3  & \textbf{71.4} & 9.5& 12.6  & 68.4 & 10.7  & 11.9  & \textbf{77.7} & 10.1  \\
    Video-Llama \cite{zhang2023video}& 17.7  & 75.4 & 38.3  & 16.4  & 71.3 & 26.5& 13.9  & \textbf{70.6} & 21.8  & 14.6  & 68.1 & 13.6  \\
    Video-Vicuna \cite{chiang2023vicuna}& 16.9  & 63.3 & 36.8 & 16.0  & 61.7 & 23.9&15.4  & 60.9 & 20.3  & 16.2  & 60.7 & 10.7  \\
    Video-LLaVA \cite{lin2023video}&20.6 & 66.3 & 40.6& 20.0  & 67.4 & 37.9& 21.7  & 65.2 & 34.2  & 20.8  & 66.1 & 30.8 \\
    V2Xformer \cite{wang2024deepaccident}&58.9 & 50.2 & 66.8& 60.6  & 53.8 & 67.4 & 65.1  & 66.7 &  68.4  & 66.3  & 57.8 & 68.9\\
    AccidentGPT \cite{wang2024accidentgpt}&60.0 & 63.7 & 68.2& 62.5  & 64.1 & 68.8& 65.0  & 67.3 & 69.1  & 69.3  & 70.1& 70.0\\
    AccidentBlip (Our) &\textbf{62.4} & 63.1 & \textbf{70.0} & \textbf{64.3}  & 64.5 & \textbf{71.6}& \textbf{67.7}  & 67.5 & \textbf{70.8}  & \textbf{71.0}  & 70.3 & \textbf{73.3} \\ 
  \bottomrule
    \end{tabularx}
    \caption{\textbf{Multi-vehicle V2V system detection results. }AccidentBlip achieves the best in both accident detection precision and accident prediction accuracy compared to other models.}
    \label{muti-ego}
    \vspace{-0.4cm}
\end{table*}

\noindent\textbf{Results of ego.} As shown in Table \ref{ego}, the video understanding models \cite{zhang2023video, chiang2023vicuna, lin2023video} demonstrate significantly higher recall among all methods but suffer from lower precision, indicating limitations in accurately understanding the surroundings. In contrast, AccidentBlip achieves \textit{SOTA} precision and accident prediction accuracy on the DeepAccident dataset \cite{wang2024deepaccident}, with a recall surpassing AccNet \cite{liao2024real}, V2Xformer \cite{wang2024deepaccident}, and AccidentGPT \cite{wang2024accidentgpt}. This highlights AccidentBlip’s superior performance in accident detection and prediction. An example shown in Fig. \ref{fig:example} illustrates that AccidentBlip can successfully detect and warn of accidents even in complex scenarios with unevenly distributed surrounding traffic.

\noindent\textbf{Results of V2X scenario. }Table \ref{ego} also represents the results in the V2X scenario. The precision, recall, and accident prediction accuracy of video understanding models decrease significantly, indicating their limitations for V2X accident detection and prediction. In contrast, AccidentBlip achieves \textit{SOTA} across all three metrics, demonstrating its superior capability for V2X accident detection and prediction.


\noindent\textbf{Results of accident prediction horizon.} To evaluate AccidentBlip's prediction horizon, we set accidents to occur at 2s, 3s, and 4s while the vehicle is in motion. As shown in Table \ref{Table:horizon}, AccidentBlip effectively predicts accidents within a 2s horizon, demonstrating strong performance. However, when the horizon is extended to 3s, the predictive ability decreases significantly as the distance increases. At 4s, the model nearly loses its predictive capability. Thus, AccidentBlip’s effective prediction limit is 4s, with accurate accident warnings within the 2s and 3s horizons.

\begin{table}[!t]
    \centering
    \resizebox{235pt}{!}{\begin{tabular}{c|ccccc}
    \toprule
        Prediction horizon & All data & 1s& 2s & 3s & 4s   \\ \midrule
        2s & 64.7 & 75.9& 30.2 & none & none  \\ 
        3s & 53.9 & 73.0& 27.8 & 23.0 & none  \\ 
        4s & 39.1 & 58.8 & 22.6& 16.5 & 12.7 
        \\ \bottomrule
    \end{tabular}}
    \caption{\textbf{Result of prediction horizon. }AccidentBlip has ability to achieve early warning before accident.}
    \label{Table:horizon}
    \vspace{-0.4cm}
\end{table}

\noindent\textbf{Result of V2X scenario.} As shown in Table \ref{muti-ego}, as system complexity increases, both the precision of detection and accident prediction accuracy of the frame-by-frame Q\&A model and the video comprehension model decline significantly. Although the video comprehension model achieves much higher recall, it does not necessarily mean it correctly identifies accidents, as it can misreport accident-free scenes. Although the multimodal detection model excels in accident prediction, its detection capability lags behind AccidentBlip. Thus, AccidentBlip achieves \textit{SOTA} performance in both accident detection and prediction tasks in the DeepAccident V2V scenario, demonstrating excellent traffic scene understanding.

\subsection{Ablation Study}

In this section, we mainly discuss the impact of different modules in AccidentBlip on model performance. Our main innovative modules are the temporal attention module of MA-former, the residual connection of $Q_{n-1}$ with $q_n$ we called $c_1$ and the residual connection of $f_n$ with $q_n$ we called $c_2$. AccidentBlip has same structure as Blip2 \cite{li2023blip} except temporal attention and the results of Blip2 are shown in Table \ref{ego}. As shown in Tables \ref{ego} and \ref{Table:abl}, AccidentBlip with residual connection has great advantages on both accident detection and prediction ability, so the contribution of temporal attention to the model's capability is obvious.

As it shown in Tab. \ref{Table:abl}, AccidentBlip's detection precision and accident prediction accuracy are significantly improved When only add residual connection $c_1$. This suggests that $c_1$ is more useful than $c_2$ for model temporal understanding ability. However, when AccidentBlip adds two residual connections as well, model's detection precision and accident prediction accuracy improve significantly than when adds only one residual connections. Thar suggests that $c_1$, $c_2$ can gain from each other and improve the model's performance together.

\begin{table}[!t]
    \centering
    \begin{tabular}{cc|ccc}
    \toprule
        $c_1$ & $c_2$& Pre. & Rec. & APA   \\ \midrule
        × & ×& 54.1 & 56.7 & 52.3  \\ 
        √ & × & 59.8 (+5.1)& 61.8 (+5.1)& 59.6 (+7.3) \\
        × & √& 54.6 (+0.5)& 57.2 (+0.5)& 54.1 (+1.8) \\ 
        √ & √ & 62.6 (+8.5)& 68.4 (+11.7)& 64.7 (+12.4)
        \\ 
        \bottomrule
    \end{tabular}
    \caption{\textbf{Result of prediction horizon. }AccidentBlip has ability to achieve early warning before accident.}
    \label{Table:abl}
    \vspace{-0.4cm}
\end{table}

\section{CONCLUSIONS}


In this paper, we present AccidentBlip, a Transformer-based agent that utilizes pure vision inputs for accident detection and prediction in V2X and V2V scenarios. By introducing the MA-former, our approach effectively processes temporal multi-view image inputs, enabling robust performance in complex traffic environments.
Experimental results on the DeepAccident dataset demonstrate that AccidentBlip achieves superior average precision in accident detection and high accuracy in accident prediction. Moreover, AccidentBlip exhibits exceptional accuracy and adaptability in both V2X and V2V systems, underscoring its potential for deployment in real-world traffic scenarios.
This work makes a significant contribution to automotive safety by providing a robust and efficient framework for accident analysis and prevention.

\addtolength{\textheight}{-0cm}   




\balance
\bibliographystyle{IEEEtran}
\bibliography{ref}  
\end{CJK}

\end{document}